\documentclass[11pt,a4paper]{article}
\usepackage[hyperref]{acl2020}
\usepackage{enumitem}
\usepackage{times}
\usepackage{latexsym}
\usepackage{graphicx}
\usepackage{booktabs}
\usepackage[export]{adjustbox}
\usepackage{dblfloatfix} 

\usepackage{microtype}

\aclfinalcopy 

\title{Low Rank Fusion based Transformers for Multimodal Sequences}

\author{ Saurav Sahay \quad Eda Okur \quad Shachi H Kumar \quad Lama Nachman \\
Intel Labs, Anticipatory Computing Lab, USA \\
\texttt{\{saurav.sahay, eda.okur, shachi.h.kumar, lama.nachman\}}\\
\texttt{@intel.com}}

\date{}

\begin{document}
\maketitle
\begin{abstract}
Our senses individually work in a coordinated fashion to express our emotional intentions. In this work, we experiment with modeling modality-specific sensory signals to attend to our latent multimodal emotional intentions and vice versa expressed via low-rank multimodal fusion and multimodal transformers. The low-rank factorization of multimodal fusion amongst the modalities helps represent approximate multiplicative latent signal interactions. Motivated by the work of~\cite{tsai2019MULT} and~\cite{Liu_2018}, we present our transformer-based cross-fusion architecture without any over-parameterization of the model. The low-rank fusion helps represent the latent signal interactions while the modality-specific attention helps focus on relevant parts of the signal. We present two methods for the Multimodal Sentiment and Emotion Recognition results on CMU-MOSEI, CMU-MOSI, and IEMOCAP datasets and show that our models have lesser parameters, train faster and perform comparably to many larger fusion-based architectures.
\end{abstract}



\section{Introduction}

The field of Emotion Understanding involves computational study of subjective elements such as sentiments, opinions, attitudes, and emotions towards other objects or persons. Subjectivity is an inherent part of emotion understanding that comes from the contextual nature of the natural phenomenon. Defining the metrics and disentangling the objective assessment of the metrics from the subjective signal makes the field quite challenging and exciting. Sentiments and Emotions are attached to the language, audio and visual modalities at different rates of expression and granularity and are useful in deriving social, psychological and behavioral insights about various entities such as movies, products, people or organizations. Emotions are defined as brief organically synchronized evaluations of major events whereas sentiments are considered as more enduring beliefs and dispositions towards objects or persons~\cite{scherer1984emotion}. The field of Emotion Understanding has rich literature with many interesting models of understanding~\cite{plutchik2001nature, ekman2009telling, posner2005circumplex}. Recent studies on tensor-based multimodal fusion explore regularizing tensor representations ~\cite{liang-etal-2019-learning} and polynomial tensor pooling~\cite{NIPS2019_9381}.

\begin{figure*}[ht!]
\centering
\includegraphics[width=\textwidth,scale=0.7]{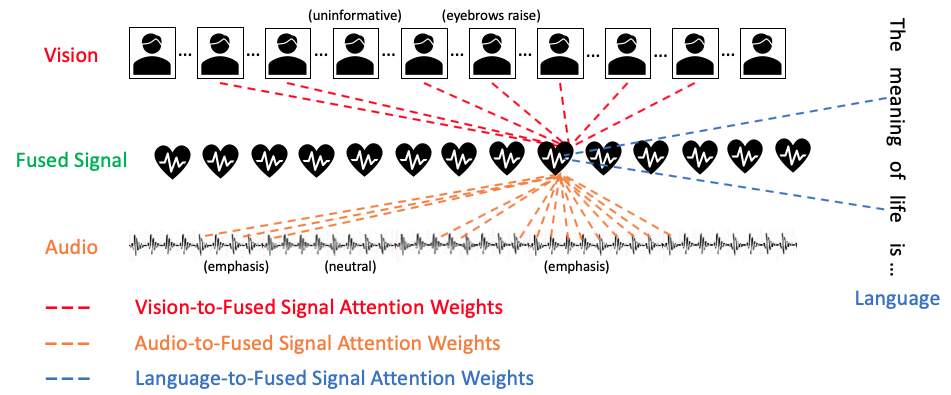}
\caption{Modality-specific Fused Attention}
    \label{sequences}
\end{figure*}

In this work, we combine ideas from~\cite{tsai2019MULT} and~\cite{Liu_2018} and explore the use of Transformer~\cite{NIPS2017_7181} based models for both aligned and unaligned signals without extensive over-parameterization of the models by using multiple modality-specific transformers. We utilize Low Rank Matrix Factorization (LMF) based fusion method for representing multimodal fusion of the modality-specific information. Our main contributions can be summarized as follows: \begin{itemize}
\item Recently proposed Multimodal Transformer (MulT) architecture~\cite{tsai2019MULT} uses at least 9 Transformer based models for cross-modal representation of language, audio and visual modalities (3 parallel modality-specific standard Transformers with self-attention and 6 parallel bimodal Transformers with cross-modal attention). These models utilize several parallel unimodal and bimodal transformers and do not capture the full trimodal signal interplay in any single transformer model in the architecture. In contrast, our method uses fewer Transformer based models and fewer parallel models for the same multimodal representation.
\item We look at two methods for leveraging the multimodal fusion into the transformer architecture. In one method (LMF-MulT), the fused multimodal signal is reinforced using attention from the 3 modalities. In the other method (Fusion-Based-CM-Attn), the individual modalities are reinforced in parallel via the fused signal.    
\end{itemize}

The ability to use unaligned sequences for modeling is advantageous since we rely on learning based methods instead of using methods that force the signal synchronization (requiring extra timing information) to mimic the coordinated nature of human multimodal language expression. The LMF method aims to capture all unimodal, bimodal and trimodal interactions amongst the modalities via approximate Tensor Fusion method. 

We develop and test our approaches on the CMU-MOSI, CMU-MOSEI, and IEMOCAP datasets as reported in~\cite{tsai2019MULT}. CMU Multimodal Opinion Sentiment and Emotion Intensity (CMU-MOSEI)~\cite{cmumoseiacl2018} is a large dataset of multimodal sentiment analysis and emotion recognition on YouTube video segments. The dataset contains more than 23,500 sentence utterance videos from more than 1000 online YouTube speakers. The dataset has several interesting properties such as being gender balanced, containing various topics and monologue videos from people with different personality traits. The videos are manually transcribed and properly punctuated. Since the dataset comprises of natural audio-visual opinionated expressions of the speakers, it provides an excellent test-bed for research in emotion and sentiment understanding. The videos are cut into continuous segments and the segments are annotated with 7 point scale sentiment labels and 4 point scale emotion categories corresponding to the Ekman's 6 basic emotion classes~\cite{10025007347}. The opinionated expressions in the segments contain visual cues, audio variations in signal as well as textual expressions showing various subtle and non-obvious interactions across the modalities for both sentiment and emotion classification. CMU-MOSI~\cite{zadeh2016multimodal} is a smaller dataset (2199 clips) of YouTube videos with sentiment annotations. IEMOCAP~\cite{iemocap} dataset consists of 10K videos with sentiment and emotion labels. We use the same setup as~\cite{tsai2019MULT} with 4 emotions (happy, sad, angry, neutral).

In Fig~\ref{sequences}, we illustrate our ideas by showing the fused signal representation attending to different parts of the unimodal sequences. There's no need to align the signals since the attention computation to different parts of the modalities acts as proxy to the multimodal sequence alignment. The fused signal is computed via Low Rank Matrix Factorization (LMF). The other model we propose uses a swapped configuration where the individual modalities attend to the fused signal in parallel.

\section{Model Description}
In this section, we describe our models and methods for Low Rank Fusion of the modalities for use with Multimodal Transformers with cross-modal attention. 

\begin{figure*}[ht!]
\centering
\includegraphics[width=\textwidth,scale=0.4]{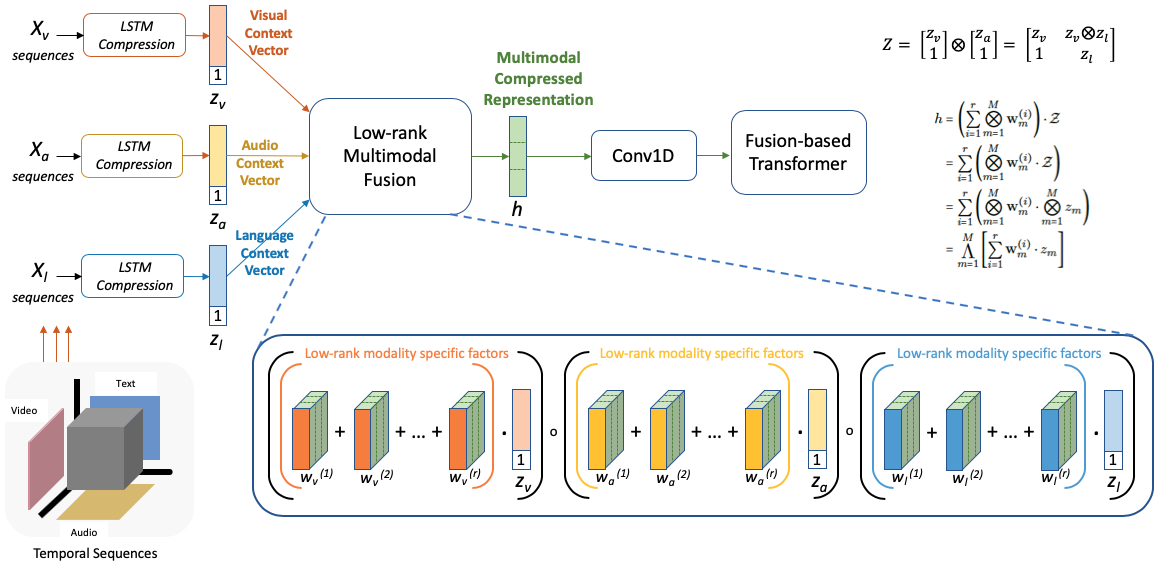}
\caption{Low Rank Matrix Factorization}
    \label{arch_lmf}
\end{figure*}

\begin{figure}[!b]
\centering
\includegraphics[scale=0.3]{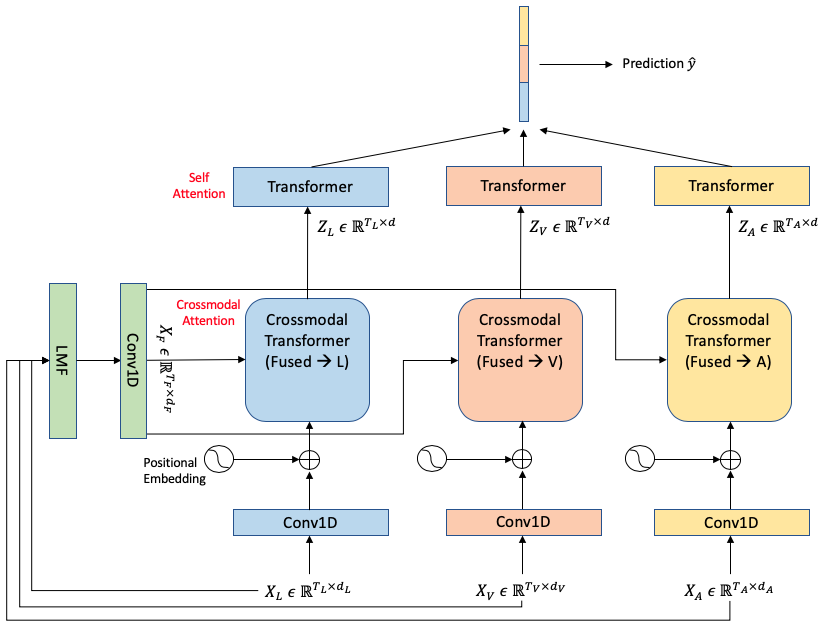}
\caption{Fused Cross-modal Transformer}
    \label{model_arch1}
\end{figure}

\subsection{Low Rank Fusion}

LMF is a Tensor Fusion method that models the unimodal, bimodal and trimodal interactions without using an expensive 3-fold Cartesian product~\cite{ZadehCPCM17} from modality-specific embeddings. Instead, the method leverages unimodal features and weights directly to approximate the full multi-tensor outer product operation. This low-rank matrix factorization operation easily extends to problems where the interaction space (feature space or number of modalities) is very large. We utilize the method as described in~\cite{Liu_2018}. Similar to the prior work, we compress the time-series information of the individual modalities using an LSTM~\cite{doi:10.1162/neco.1997.9.8.1735} and extract the hidden state context vector for modality-specific fusion. We depict the LMF method in Fig~\ref{arch_lmf} similar to the illustration in~\cite{Liu_2018}. This shows how the unimodal tensor sequences are appended with 1s before taking the outer product to be equivalent to the tensor representation that captures the unimodal and multimodal interaction information explicitly (top right of Fig~\ref{arch_lmf}). As shown, the compressed representation (h) is computed using batch matrix multiplications of the low-rank modality-specific factors and the appended modality representations. All the low-rank products are further multiplied together to get the fused vector.  

\begin{figure}[!b]
\centering
\includegraphics[scale=0.3]{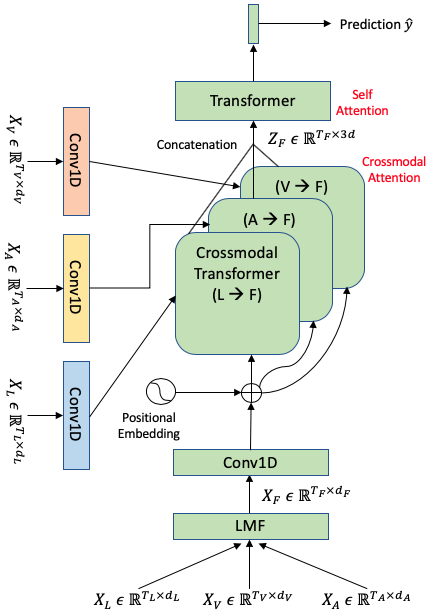}
\caption{Low Rank Fusion Transformer}
    \label{model_arch2}
\end{figure}

\begin{table*}[t!]
\centering
\small
\begin{tabular}{|c||ccccc|}
\hline
Metric & Acc$_7^h$ & Acc$_2^h$ & F1$^h$ & MAE$^l$ & Corr$^h$ \\
\hline\hline
\multicolumn{6}{|c|}{(Aligned) CMU-MOSI Sentiment} \\
\hline\hline
LF-LSTM (pub) & 35.3 & 76.8 & 76.7 & 1.015 & 0.625 \\
MulT~\cite{tsai2019MULT} (pub) & 40.0 & 83.0 & 82.8 & 0.871 & 0.698 \\
\hline
MulT~\cite{tsai2019MULT} (our run) & \textbf{33.1} & \textbf{78.5} & \textbf{78.4} & \textbf{0.991} & \textbf{0.676} \\
Fusion-Based-CM-Attn-MulT (ours) & 32.9 & 77.0 & 76.9 & 1.017 & 0.636 \\
LMF-MulT (ours) & 32.4 & 77.9 & 77.9 & 1.016 & 0.647 \\
\hline\hline
\multicolumn{6}{|c|}{(Unaligned) CMU-MOSI Sentiment} \\
\hline\hline
LF-LSTM (pub) & 33.7 & 77.6 & 77.8 & 0.988 & 0.624 \\
MulT~\cite{tsai2019MULT} (pub) & 39.1 & 81.1 & 81.0 & 0.889 & 0.686 \\
\hline
MulT~\cite{tsai2019MULT} (our run) & 34.3 & \textbf{80.3} & \textbf{80.4} & 1.008 & 0.645 \\
Fusion-Based-CM-Attn-MulT (ours) & \textbf{34.4} & 76.8 & 76.8 & 1.003 & 0.640 \\
LMF-MulT (ours) & 34.0 & 78.5 & 78.5 & \textbf{0.957} & \textbf{0.681} \\
\hline
\end{tabular}
\caption{Performance Results for Multimodal Sentiment Analysis on CMU-MOSI dataset with aligned and unaligned multimodal sequences.}
\label{MOSI} 
\end{table*}

\begin{table*}[t!]
\centering
\small
\begin{tabular}{|c||ccccc|}
\hline
Metric & Acc$_7^h$ & Acc$_2^h$ & F1$^h$ & MAE$^l$ & Corr$^h$ \\
\hline\hline
\multicolumn{6}{|c|}{(Aligned) CMU-MOSEI Sentiment} \\
\hline\hline
LF-LSTM (pub) & 48.8 & 80.6 & 80.6 & 0.619 & 0.659 \\
MulT~\cite{tsai2019MULT} (pub) & 51.8 & 82.5 & 82.3 & 0.580 & 0.703 \\
\hline
MulT~\cite{tsai2019MULT} (our run) & 49.3 & \textbf{80.5} & \textbf{81.1} & 0.625 & 0.663 \\
Fusion-Based-CM-Attn-MulT (ours) & 49.6 & 79.9 & 80.7 & \textbf{0.616} & \textbf{0.673} \\
LMF-MulT (ours) & \textbf{50.2} & 80.3 & 80.3 & \textbf{0.616} & 0.662 \\
\hline\hline
\multicolumn{6}{|c|}{(Unaligned) CMU-MOSEI Sentiment} \\
\hline\hline
LF-LSTM (pub) & 48.8 & 77.5 & 78.2 & 0.624 & 0.656 \\
MulT~\cite{tsai2019MULT} (pub) & 50.7 & 81.6 & 81.6 & 0.591 & 0.694 \\
\hline
MulT~\cite{tsai2019MULT} (our run) & \textbf{50.4} & 80.7 & 80.6 & 0.617 & \textbf{0.677} \\
Fusion-Based-CM-Attn-MulT (ours) & 49.3 & 79.4 & 79.2 & \textbf{0.613} & 0.674 \\
LMF-MulT (ours) & 49.3 & \textbf{80.8} & \textbf{81.3} & 0.620 & 0.668 \\
\hline
\end{tabular}
\caption{Performance Results for Multimodal Sentiment Analysis on larger-scale CMU-MOSEI dataset with aligned and unaligned multimodal sequences.}
\label{MOSEI} 
\end{table*}

\subsection{Multimodal Transformer}
We build up on the Transformers~\cite{NIPS2017_7181} based sequence encoding and utilize the ideas from~\cite{tsai2019MULT} for multiple cross-modal attention blocks followed by self-attention for encoding multimodal sequences for classification. While the earlier work focuses on latent adaptation of one modality to another, we focus on  adaptation of the latent multimodal signal itself using single-head cross-modal attention to individual modalities. This helps us reduce the excessive parameterization of the models by using all combinations of modality to modality cross-modal attention for each modality. Instead, we only utilize a linear number of cross-modal attention for each modality and the fused signal representation. We add Temporal Convolutions after the LMF operation to ensure that the input sequences have a sufficient awareness of the neighboring elements. We show the overall architecture of our two proposed models in Fig~\ref{model_arch1} and Fig~\ref{model_arch2}. In Fig~\ref{model_arch1}, we show the fused multimodal signal representation after a temporal convolution to enrich the individual modalities via cross-modal transformer attention. In Fig~\ref{model_arch2}, we show the architecture with the least number of Transformer layers where the individual modalities attend to the fused convoluted multimodal signal. 

\begin{table*}[t!]
\centering
\small
\begin{tabular}{|c||cccccccc|}
\hline
Emotion & \multicolumn{2}{c}{Happy} & \multicolumn{2}{c}{Sad} & \multicolumn{2}{c}{Angry} & \multicolumn{2}{c|}{Neutral} \\
Metric & Acc$^h$ & F1$^h$ & Acc$^h$ & F1$^h$ & Acc$^h$ & F1$^h$ & Acc$^h$ & F1$^h$ \\
\hline\hline
\multicolumn{9}{|c|}{(Aligned) IEMOCAP Emotions} \\
\hline\hline
LF-LSTM (pub) & 85.1 & 86.3 & 78.9 & 81.7 & 84.7 & 83.0 & 67.1 & 67.6 \\
MulT~\cite{tsai2019MULT} (pub) & 90.7 & 88.6 & 86.7 & 86.0 & 87.4 & 87.0 & 72.4 & 70.7 \\
\hline
MulT~\cite{tsai2019MULT} (our run) & \textbf{86.4} & 82.9 & 82.3 & 82.4 & 85.3 & 85.8 & \textbf{71.2} & 70.0 \\
Fusion-Based-CM-Attn-MulT (ours) & 85.6 & 83.7 & 83.6 & \textbf{83.7} & 84.6 & 85.0 & 70.4 & 69.9 \\
LMF-MulT (ours) & 85.3 & \textbf{84.1} & \textbf{84.1} & 83.4 & \textbf{85.7} & \textbf{86.2} & \textbf{71.2} & \textbf{70.8} \\
\hline\hline
\multicolumn{9}{|c|}{(Unaligned) IEMOCAP Emotions} \\
\hline\hline
LF-LSTM (pub) & 72.5 & 71.8 & 72.9 & 70.4 & 68.6 & 67.9 & 59.6 & 56.2 \\
MulT~\cite{tsai2019MULT} (pub) & 84.8 & 81.9 & 77.7 & 74.1 & 73.9 & 70.2 & 62.5 & 59.7 \\
\hline
MulT~\cite{tsai2019MULT} (our run) & \textbf{85.6} & \textbf{79.0} & \textbf{79.4} & \textbf{70.3} & \textbf{75.8} & \textbf{65.4} & 59.2 & 44.0 \\
Fusion-Based-CM-Attn-MulT (ours) & \textbf{85.6} & \textbf{79.0} & \textbf{79.4} & \textbf{70.3} & \textbf{75.8} & \textbf{65.4} & \textbf{59.3} & \textbf{44.2}\\
LMF-MulT (ours) & \textbf{85.6} & \textbf{79.0} & \textbf{79.4} & \textbf{70.3} & \textbf{75.8} & \textbf{65.4} & 59.2 & 44.0 \\
\hline
\end{tabular}
\caption{Performance Results for Multimodal Emotion Recognition on IEMOCAP dataset with aligned and unaligned multimodal sequences.}
\label{IEMOCAP} 
\end{table*}

\begin{table*}[t!]
\centering
\small
\begin{tabular}{ |c||c|c|c|c|c|c| } 
 \hline
 Dataset & \multicolumn{2}{c|}{CMU-MOSI} & \multicolumn{2}{c|}{CMU-MOSEI} & \multicolumn{2}{c|}{IEMOCAP} \\
 \hline
 Model & Aligned & Unaligned & Aligned & Unaligned & Aligned & Unaligned \\ [0.5ex] 
 \hline\hline
 MulT~\cite{tsai2019MULT} & 18.87 & 19.25 & 191.40 & 216.32 & 36.20 & 37.93 \\ 
 Fusion-Based-CM-Attn (ours) & 14.53 & 15.80 & 140.95 & 175.68 & 26.10 & 29.16 \\ 
 LMF-MulT (ours) & \textbf{11.01} & \textbf{12.03} & \textbf{106.15} & \textbf{137.35} & \textbf{20.57} & \textbf{23.53} \\ 
 \hline
\end{tabular}
\caption{Average Time/Epoch (sec)}
\label{time_epoch}
\end{table*}

\begin{table*}[h!]
\centering
\small
\begin{tabular}{ |c||c|c|c| } 
 \hline
 Dataset & CMU-MOSI & CMU-MOSEI & IEMOCAP \\
 \hline\hline
 MulT~\cite{tsai2019MULT} & 1071211 & 1073731 & 1074998 \\ 
 Fusion-Based-CM-Attn (ours) & \textbf{512121} & \textbf{531441} & \textbf{532078} \\ 
 LMF-MulT (ours) & 836121 & 855441 & 856078 \\ 
 \hline
\end{tabular}
\caption{Number of Model Parameters}
\label{model_params}
\end{table*}


\section{Experiments}

We present our early experiments to evaluate the performance of proposed models on the standard multimodal datasets used by~\cite{tsai2019MULT}\footnote{We have built this work up on the code-base released for MulT~\cite{tsai2019MULT} at~\url{https://github.com/yaohungt/Multimodal-Transformer}}. We run our models on CMU-MOSI, CMU-MOSEI, and IEMOCAP datasets and present the results for the proposed LMF-MulT and Fusion-Based-CM-Attn-MulT models. Late Fusion (LF) LSTM  is a common baseline for all datasets with reported results (pub) together with MulT in~\cite{tsai2019MULT}. We include the results we obtain (our run) for the MulT model for a direct comparison\footnote{In this work, we have not focused on the further hyper-parameter tuning of our models.}.
Table~\ref{MOSI}, Table~\ref{MOSEI}, and Table~\ref{IEMOCAP} show the performance of various models on the sentiment analysis and emotion classification datasets. We do not observe any trend suggesting that our methods can achieve better accuracies or F1-scores than the original MulT method~\cite{tsai2019MULT}. However, we do note that on some occasions, our methods can achieve higher results than the MulT model, in both aligned (see LMF-MulT results for IEMOCAP in Table~\ref{IEMOCAP}) and unaligned (see LMF-MulT results for CMU-MOSEI in Table~\ref{MOSEI}) case. We plan to do an exhaustive grid search over the hyper-parameters to understand if our methods can learn to classify the multimodal signal better than the original competitive method. Although the results are comparable, below are the advantages of using our methods:
\begin{itemize}
\item Our LMF-MulT model does not use multiple parallel self-attention transformers for the different modalities and it uses least number of transformers compared to the other two models. Given the same training infrastructure and resources, we observe a consistent speedup in training with this method. See Table~\ref{time_epoch} for average time per epoch in seconds measured with fixed batch sizes for all three models.
\item As summarized in Table ~\ref{model_params}, we observe that our models use lesser number of trainable parameters compared to the MulT model, and yet achieve similar performance.
\end{itemize}

\section{Conclusion}
In this paper, we present our early investigations towards utilizing Low Rank representations of the multimodal sequences for usage in multimodal transformers with cross-modal attention to the fused signal or the modalities. Our methods build up on the~\cite{tsai2019MULT} work and apply transformers to fused multimodal signal that aim to capture all inter-modal signals via the Low Rank Matrix Factorization~\cite{Liu_2018}. This method is applicable to both aligned and unaligned sequences. Our methods train faster and use fewer parameters to learn classifiers with similar SOTA performance. We are exploring methods to compress the temporal sequences without using the hidden state context vectors from LSTMs that lose the temporal information. We recover the temporal information with a Convolution layer. We believe these models can be deployed in low resource settings with further optimizations. We are also interested in using richer features for the audio, text, and the vision pipeline for other use-cases where we can utilize more resources. \\



\bibliography{acl2020}

\begin{thebibliography}{15}
\expandafter\ifx\csname natexlab\endcsname\relax\def\natexlab#1{#1}\fi

\bibitem[{Busso et~al.(2008)Busso, Bulut, Lee, Kazemzadeh, Mower, Kim, Chang,
  Lee, and Narayanan}]{iemocap}
Carlos Busso, Murtaza Bulut, Chi-Chun Lee, Abe Kazemzadeh, Emily Mower, Samuel
  Kim, Jeannette~N. Chang, Sungbok Lee, and Shrikanth~S. Narayanan. 2008.
\newblock \href {https://doi.org/10.1007/s10579-008-9076-6} {Iemocap:
  interactive emotional dyadic motion capture database}.
\newblock \emph{Language Resources and Evaluation}, 42(4):335.

\bibitem[{Ekman(2002)}]{10025007347}
Paul Ekman. 2002.
\newblock \href {https://ci.nii.ac.jp/naid/10025007347/en/} {Facial action
  coding system (facs)}.
\newblock \emph{A Human Face}.

\bibitem[{Ekman(2009)}]{ekman2009telling}
Paul Ekman. 2009.
\newblock \emph{Telling Lies: Clues to Deceit in the Marketplace, Politics, and
  Marriage (Revised Edition)}.
\newblock WW Norton \& Company.

\bibitem[{Hochreiter and Schmidhuber(1997)}]{doi:10.1162/neco.1997.9.8.1735}
Sepp Hochreiter and Jürgen Schmidhuber. 1997.
\newblock \href {https://doi.org/10.1162/neco.1997.9.8.1735} {Long short-term
  memory}.
\newblock \emph{Neural Computation}, 9(8):1735--1780.

\bibitem[{Hou et~al.(2019)Hou, Tang, Zhang, Kong, and Zhao}]{NIPS2019_9381}
Ming Hou, Jiajia Tang, Jianhai Zhang, Wanzeng Kong, and Qibin Zhao. 2019.
\newblock \href
  {http://papers.nips.cc/paper/9381-deep-multimodal-multilinear-fusion-with-high-order-polynomial-pooling.pdf}
  {Deep multimodal multilinear fusion with high-order polynomial pooling}.
\newblock In \emph{Advances in Neural Information Processing Systems 32}, pages
  12136--12145. Curran Associates, Inc.

\bibitem[{Liang et~al.(2019)Liang, Liu, Tsai, Zhao, Salakhutdinov, and
  Morency}]{liang-etal-2019-learning}
Paul~Pu Liang, Zhun Liu, Yao-Hung~Hubert Tsai, Qibin Zhao, Ruslan
  Salakhutdinov, and Louis-Philippe Morency. 2019.
\newblock \href {https://doi.org/10.18653/v1/P19-1152} {Learning
  representations from imperfect time series data via tensor rank
  regularization}.
\newblock In \emph{Proceedings of the 57th Annual Meeting of the Association
  for Computational Linguistics}, pages 1569--1576, Florence, Italy.
  Association for Computational Linguistics.

\bibitem[{Liu et~al.(2018)Liu, Shen, Lakshminarasimhan, Liang, Bagher~Zadeh,
  and Morency}]{Liu_2018}
Zhun Liu, Ying Shen, Varun~Bharadhwaj Lakshminarasimhan, Paul~Pu Liang, AmirAli
  Bagher~Zadeh, and Louis-Philippe Morency. 2018.
\newblock \href {https://doi.org/10.18653/v1/p18-1209} {Efficient low-rank
  multimodal fusion with modality-specific factors}.
\newblock \emph{Proceedings of the 56th Annual Meeting of the Association for
  Computational Linguistics (Volume 1: Long Papers)}.

\bibitem[{Plutchik(2001)}]{plutchik2001nature}
Robert Plutchik. 2001.
\newblock The nature of emotions human emotions have deep evolutionary roots, a
  fact that may explain their complexity and provide tools for clinical
  practice.
\newblock \emph{American Scientist}.

\bibitem[{Posner et~al.(2005)Posner, Russell, and
  Peterson}]{posner2005circumplex}
Jonathan Posner, James~A Russell, and Bradley~S Peterson. 2005.
\newblock The circumplex model of affect: An integrative approach to affective
  neuroscience, cognitive development, and psychopathology.
\newblock \emph{Development and psychopathology}, 17(03):715--734.

\bibitem[{Scherer(1984)}]{scherer1984emotion}
Klaus~R Scherer. 1984.
\newblock Emotion as a multicomponent process: A model and some cross-cultural
  data.
\newblock \emph{Review of Personality \& Social Psychology}.

\bibitem[{Tsai et~al.(2019)Tsai, Bai, Liang, Kolter, Morency, and
  Salakhutdinov}]{tsai2019MULT}
Yao-Hung~Hubert Tsai, Shaojie Bai, Paul~Pu Liang, J.~Zico Kolter,
  Louis-Philippe Morency, and Ruslan Salakhutdinov. 2019.
\newblock Multimodal transformer for unaligned multimodal language sequences.
\newblock In \emph{Proceedings of the 57th Annual Meeting of the Association
  for Computational Linguistics (Volume 1: Long Papers)}, Florence, Italy.
  Association for Computational Linguistics.

\bibitem[{Vaswani et~al.(2017)Vaswani, Shazeer, Parmar, Uszkoreit, Jones,
  Gomez, Kaiser, and Polosukhin}]{NIPS2017_7181}
Ashish Vaswani, Noam Shazeer, Niki Parmar, Jakob Uszkoreit, Llion Jones,
  Aidan~N Gomez, \L~ukasz Kaiser, and Illia Polosukhin. 2017.
\newblock \href
  {http://papers.nips.cc/paper/7181-attention-is-all-you-need.pdf} {Attention
  is all you need}.
\newblock In I.~Guyon, U.~V. Luxburg, S.~Bengio, H.~Wallach, R.~Fergus,
  S.~Vishwanathan, and R.~Garnett, editors, \emph{Advances in Neural
  Information Processing Systems 30}, pages 5998--6008. Curran Associates, Inc.

\bibitem[{Zadeh et~al.(2017)Zadeh, Chen, Poria, Cambria, and
  Morency}]{ZadehCPCM17}
Amir Zadeh, Minghai Chen, Soujanya Poria, Erik Cambria, and Louis{-}Philippe
  Morency. 2017.
\newblock \href {http://arxiv.org/abs/1707.07250} {Tensor fusion network for
  multimodal sentiment analysis}.
\newblock \emph{CoRR}, abs/1707.07250.

\bibitem[{Zadeh et~al.(2018)Zadeh, Liang, Vanbriesen, Poria, Cambria, Chen, and
  Morency}]{cmumoseiacl2018}
Amir Zadeh, Paul~Pu Liang, Jon Vanbriesen, Soujanya Poria, Erik Cambria,
  Minghai Chen, and Louis-Philippe Morency. 2018.
\newblock Multimodal language analysis in the wild: Cmu-mosei dataset and
  interpretable dynamic fusion graph.
\newblock In \emph{Association for Computational Linguistics (ACL)}.

\bibitem[{Zadeh et~al.(2016)Zadeh, Zellers, Pincus, and
  Morency}]{zadeh2016multimodal}
Amir Zadeh, Rowan Zellers, Eli Pincus, and Louis-Philippe Morency. 2016.
\newblock Multimodal sentiment intensity analysis in videos: Facial gestures
  and verbal messages.
\newblock \emph{IEEE Intelligent Systems}, 31(6):82--88.

\end{thebibliography}
\bibliographystyle{acl_natbib}

\end{document}